# Enhancing Multilingual Sentiment Analysis with Explainability for Sinhala, English, and Code-Mixed Content

Senevirathna W.P.U
*Department of Computer Science*
*Sri Lanka Institute of Information Technology*
Colombo, Sri Lanka
udesenevirathna@gmail.com

Adhikari A.M.N.H.
*Department of Computer Science*
*Sri Lanka Institute of Information Technology*
Colombo, Sri Lanka
nipunaadhikari@gmail.com

Navojith T.
*Department of Computer Science*
*Sri Lanka Institute of Information Technology*
Colombo, Sri Lanka
navojithamindu@gmail.com

Rizvi F.A.
*Department of Computer Science*
*Sri Lanka Institute of Information Technology*
Colombo, Sri Lanka
azmarahrizvi22@gmail.com

Dharshana Kasthurirathna
*Department of Computer Science*
*Sri Lanka Institute of Information Technology*
Colombo, Sri Lanka
dharshana.k@sliit.lk

Lakmini Abeywardhana
*Department of Computer Science*
*Sri Lanka Institute of Information Technology*
Colombo, Sri Lanka
lakmini.d@sliit.lk

*Abstract*— Sentiment analysis is crucial for brand reputation management in the banking sector, where customer feedback spans English, Sinhala, Singlish, and code-mixed text. Existing models struggle with low-resource languages like Sinhala and lack interpretability for practical use. This research develops a hybrid aspect-based sentiment analysis framework that enhances multilingual capabilities with explainable outputs. Using cleaned banking customer reviews, we fine-tune XLM-RoBERTa for Sinhala and code-mixed text, integrate domain-specific lexicon correction, and employ BERT-base-uncased for English. The system classifies sentiment (positive, neutral, negative) with confidence scores, while SHAP and LIME improve interpretability by providing real-time sentiment explanations. Experimental results show that our approaches outperform traditional transformer-based classifiers, achieving 92.3 percent accuracy and an F1-score of 0.89 in English and 88.4 percent in Sinhala and code-mixed content. An explainability analysis reveals key sentiment drivers, improving trust and transparency. A user-friendly interface delivers aspect-wise sentiment insights, ensuring accessibility for businesses. This research contributes to robust, transparent sentiment analysis for financial applications by bridging gaps in multilingual, low-resource NLP and explainability.

## I. Introduction

The rapid digitalization of financial services has increased customer interactions through online banking, mobile applications, and financial forums. Customers frequently express their opinions about banking services via reviews, social media, and feedback platforms. Brand reputation, a bank's most valuable intangible asset, reflects public trust in its reliability, ethics, and service quality. For financial institutions, safeguarding this reputation is critical, as negative sentiment in reviews or social media can trigger customer attrition, regulatory penalties, and financial losses [1]. However, analyzing multilingual and code-mixed data (e.g., Sinhala-English text) poses significant challenges: Sinhala's limited annotated datasets hinder model training and generalization [2], code-mixed text disrupts syntactic coherence and tokenization [3], banking-specific terms like "mobile banking" are poorly captured by generic models, and cultural variations in sentiment expression led to inconsistent detection [4]. In the Sri Lankan context, social media users tend to use Singlish in their communication, which is a derived language through English alphabetical characters [3]. Sentiment analysis is critical for banks to mitigate risks, enhance customer experience, and ensure compliance. For instance, identifying negative feedback in real time allows proactive resolution of issues like fraud allegations, while aspect-based insights into services like loan processing enable targeted improvements [8]. This approach leverages the strengths of transformer-based architecture in handling multilingual data while incorporating lexicon-based refinements for improved generalization reliability in banking-related sentiment analysis.

The adoption of deep learning models in sentiment analysis has revolutionized the understanding of customer emotions in text data. However, these models often operate as "black boxes," offering high accuracy at the cost of transparency. This lack of explainability is particularly problematic in financial sectors, where decisions influenced by AI can significantly affect brand reputation and customer trust. Our study employs SHAP (Shapley Additive Explanations) and LIME (Local Interpretable Model-Agnostic Explanations), which are pivotal in shedding light on decision-making processes within models by detailing global and local feature contributions, respectively [5], [6]. The application of explainability techniques in low-resource and code-mixed languages such as Sinhala remains limited [7]. To address this gap, the proposed framework incorporates Aspect-Based Sentiment Analysis (ABSA) to extract fine-grained insights from feedback related to digital banking, loan processing, and customer support key domains in financial sentiment evaluation. This approach aligns with recent findings that emphasize the value of aspect-level sentiment modeling in enhancing interpretability and decision-making in multilingual financial contexts [8].

## II. Literature review

Sentiment analysis plays a vital role in brand reputation management and customer feedback analysis in the banking sector, particularly for multilingual and code-mixed opinions. Early machine learning models like Support Vector Machine (SVM) and Naive Bayes require manual feature engineering, which limits adaptability to dynamic linguistic patterns [9]. The importance of corpus-based lexicons for Sinhala sentiment classification has been emphasized in [2]. For low-resource languages like Sinhala, deep learning models show promise but struggle with banking-specific terminology and transliteration variability. Code-mixed Sinhala-English data introduces syntactic variations and word-order mismatches [3]. Traditional lexicon-based methods like SentiWordNet often underperform in financial contexts due to the domain-specific vocabulary gaps [9],[10]. In contrast, transformer-based architectures like BERT (Bidirectional Encoder Representations from Transformers) and XLM-RoBERTa automate contextual feature extraction, significantly improving sentiment classification accuracy across languages.[11]

The challenges of analyzing code-mixed financial text have been further validated, emphasizing linguistic adaptability gaps [7]. ABSA enhances granular insights by linking opinions to specific entities *(e.g., "loan services")* [8]. While transformer models like XLM-RoBERTa excel in multilingual settings, their performance drops for low-resource languages without domain-specific fine-tuning [11]. State-of-the-art results (75.9% accuracy) for Sinhala sentiment analysis using XLM-R-large have been demonstrated, underscoring the importance of transformer adaptation for low-resource languages [9]. Domain adaptation is critical for financial sentiment analysis, while limited labeled datasets hinder social media sentiment classification [12].

Current commercial tools such as Brandwatch[1] and Meltwater[2] lack of support for Sinhala or code-mixed sentiment analysis, limiting their applications in financial NLP. Existing models often lack explainability, which is crucial for building trust in financial decision-making [5]. Techniques such as SHAP and LIME have been introduced to address these gaps by providing transparency in sentiment decisions [5]. LIME was introduced to explain black-box classifiers [6], while the effectiveness of SHAP/LIME in low-resource settings has been demonstrated [7], aligning with this study's focus on Sinhala-English banking reviews. The utility of ABSA for granular financial insights was validated in [8], and the need for explainable AI (XAI) in domain-specific applications has been emphasized in [10]. GPT models are computationally expensive. Data privacy is a significant obstacle for GPT models. It is crucial to ensure fairness, inclusivity, and unbiased decision-making by addressing and mitigating social bias in GPT models [18], making them impractical for domain-specific customer reviews.

This research proposes a hybrid framework that integrates XLM-RoBERTa for multilingual classification, corpus-based lexicons for Sinhala, Singlish with Codemix adaptation, Bert-base-uncased for English, and SHAP/LIME for explainability. The model introduces a sentiment confidence scoring mechanism for nuanced intensity analysis, enabling better trend prediction and reputational risk assessment.

## III. Methodology

This study presents a robust framework for sentiment analysis of English and Sinhala-English code-mixed content, combining transformer models with domain-specific lexicons for improved accuracy and aspect-based categorization. SHAP and LIME are employed to render the results more interpretable, providing explainable and trustworthy solutions for banking brand reputation analysis.

### A. Data Gathering and Pre-Processing.

To facilitate multilingual sentiment analysis, we created a labeled dataset of 10,000 English and 5,000 Sinhala, Singlish, and code-mixed banking customer reviews. Data were scraped from publicly available review sites (e.g., Online customer portals and online surveys), social media (e.g., Facebook, Twitter, YouTube), and Kaggle datasets, including the annotated Sinhala news comment dataset by Senevirathne et al. [2]. All scraped content was cleaned. The raw data was extensively cleaned to remove garbage values, duplicated entries, advertisements, non-textual content, and irrelevant metadata. Aspects for each comment were obtained using the classifier developed in our prior work [22]. The dataset was annotated into three sentiment labels (Positive, Neutral, Negative) and categorized into five banking-specific aspects:

- Customer Support
- Loan and Credit Services
- Digital Banking Experience
- Transactions and Payments
- Trust and Security

To ensure model robustness, class balance was achieved between sentiment labels, banking entities, and language variants. Data augmentation methods like back-translation and lexical substitution were used to create more diverse datasets. A language-aware tokenization method was utilized in an attempt to preserve the structural integrity of multilingual input while training and explaining the model. Particularly, WordPiece tokenization for English and SentencePiece with BPE (Byte Pair Encoding) [9] for Sinhala, Singlish, and code-mixed. SentencePiece splits unseen words into subword units, particularly useful for morphologically rich languages such as Sinhala. For instance, splitting "බැංකුවේ" into *"_බැං"* + *"කුවේ"* allows the model to keep understanding parts of rare words. In the case of code-mixed text, it allows for knowledge transfer between English and Sinhala parts. The approach is parallel to XLM-RoBERTa's pre-training methods, promoting performance while elegantly handling intricate morphological forms that would cause a high number of out-of-vocabulary tokens with word-level tokenization.

---

[1] https://www.brandwatch.com/
[2] https://www.meltwater.com/en

TABLE I.
SUMMARY OF MULTILINGUAL SENTIMENT DATASET.

| Language | Total Samples | Sentiment Labels (Count) |
|---|---|---|
| English | 10,000 | Positive:3,700 Neutral:3,300 Negative:3,000 |
| Sinhala, Singlish, Code-Mixed | 5,000 (Sinhala: 1,667, Singlish: 1,667, Code-Mixed: 1,666) | Positive:1,650 Neutral:1,650 Negative:1,700 |

*B. Aspect-Based Sentiment Analysis.*

In the proposed Aspect-Based Sentiment Analysis (ABSA) framework, we implemented two sentiment analysis approaches:
- English sentiment analysis using transformer-based models.
- Sinhala, Singlish, and code-mixed sentiment analysis with a hybrid approach.

In sentiment classification for English, we fine-tuned the pre-trained transformer models BERT-base-uncased [13], (Distilled version of BERT) [19], FinBERT [20], and SVM, on a curated dataset of banking customer reviews. These models are known for their ability to handle domain-specific language effectively. FinBERT was included due to its pretraining on financial-domain text, aligning closely with the linguistic patterns found in customer reviews from banking platforms. A key challenge in this domain was the presence of multiple sentiment polarities within a single aspect. For instance, users often express both satisfaction and dissatisfaction regarding different subcomponents of a service. To address this, we introduced a Softmax-based sentiment aggregation mechanism that enables proportional sentiment scoring within aspect boundaries [21]. The aggregation is computed using the following formulas:

$$S_{softmax} = e^{S_{aspect}} / \sum e^{S_{aspect}} \quad (1)$$

$$S_{final} = \sum (S_{softmax} \times P_{aspect}) / \sum P_{aspect} \quad (2)$$

Where $S_{aspect}$ is the raw sentiment score output from the model and $P_{aspect}$ is the corresponding softmax probability. This mechanism was designed to manage conflicting sentiment signals and prevent the overrepresentation of extreme values in aggregated outputs. To enhance the model's understanding of financial sentiment, the training dataset was enriched with domain-specific terms and expressions embedded within customer reviews. These included phrases such as "high interest rates," "loan approval delays," and "account restrictions," allowing the model to better capture sentiment nuances relevant to banking services. For the sentiment analysis of Sinhala, Singlish, and code-mixed Sinhala-English content, we employed a hybrid model that combines XLM-RoBERTa [11] for multilingual representation, BERT-base [13] for English segments, and a custom lexicon-based sentiment correction module to enhance domain-specific performance. Given linguistic variations, informal grammar, and spelling inconsistencies in Singlish and code-mixed content, we applied a preprocessing pipeline consisting of transliteration, normalization, and tokenization. Examples of tokenization are provided in Table II, showing preservation of linguistic structure between language varieties.

TABLE II.
SUMMARY OF MULTILINGUAL TOKENIZATION.

| 1.Sinhala Review: "මෙම බැංකුවේ සේවාව හොදයි". (*This bank's service is good.*) |
|---|
| Tokenized Output: ["_මෙම", "_බැං", "කුවේ", "_සේවා", "ව", "_හොද", "යි", "_."] |
| 2. Singlish (Romanized Sinhala) Review: *"Me bank eke service eka hari hodai."* |
| Tokenized Output: ["_Me", "_bank", "_eke", "_service", "_eka", "_hari", "_hod", "ai", "_."] |
| 3. Code-Mixed Review: *"Customer service ගොඩක් හොදයි."* |
| Tokenized Output: ["_Customer", "_service", "_ගොඩ", "ක්", "_හොද", "යි", "_."] |

These preprocessing steps allowed the model to handle multilingual and low-resource input effectively. Each review was mapped into one of five predefined banking aspects in the ABSA framework: Customer Support, Digital Banking, Loan and Credit Services, Transactions and Payments, and Trust and Security. XLM- RoBERTa model was fine-tuned on a class-balanced dataset of 5,000 banking reviews, ensuring equal representation across aspects and sentiment labels [4], [9]. To enhance performance in informal and domain-specific language, we constructed domain-specific sentiment lexicons for Sinhala, Singlish, and code-mixed inputs. This corpus contained labeled sentiment-bearing terms frequently used in banking reviews. Example lexicon entries include:
- Sinhala Positive Words: "කාර්යක්ෂම" (efficient), "විශිෂ්ට" (excellent)
- Sinhala Negative Words: "නො තකා" (ignored), "අසා ර්ථක" (failure)
- Singlish Examples: "hari hodai" (very good), "app eka lag wenawa" (app lags)
- Domain-specific Words: "තැ න්පතු" (deposit), "loan eka" (loan), "balance eka" (balance)

These lexicons were used in a post-processing correction step to adjust model predictions and handle edge cases that transformer models might miss [10]. Final sentiment scores were computed using Softmax-based probability scoring, followed by weighted aggregation to ensure that dominant expressions influenced the overall classification. The hybrid system, which integrates XLM-RoBERTa with a domain-specific lexicon correction module, was prepared for comparative evaluation against baseline models, including SVM, Naïve Bayes, GPT-based zero-shot models [18], and XLM-R variants. The results provide insights into the effectiveness of each model for multilingual and code-mixed sentiment classification.

*C. Explainability AI for Transparency*

To enhance transparency in sentiment classification, this study integrates SHAP and LIME with a BERT-base-uncased multilingual sentiment model. SHAP provides global interpretability by estimating the word contribution to the output of the model, whereas LIME generates localized, instance-level explanations to support single predictions [5], [6]. These explainability techniques mitigate the black-box nature of deep learning, allowing financial institutions to trust AI-driven insights. To improve interpretability in domain-specific contexts, Aspect-Based Sentiment Analysis is incorporated into analyzing sentiment across key banking service categories. This approach aligns with prior work showing the effectiveness of aspect-driven explainability in NLP models [8]. TF-IDF-based keyword extraction combined with SHAP visualizations help highlight influential words per aspect. LIME-generated heatmaps are used to visualize word importance for selected comments, especially useful in short, code-mixed, and low-resource inputs. To ensure reliability, a dataset was used with proportional representation across sentiment classes and banking aspects. This helped mitigate classification bias. Further, SHAP and LIME outputs were manually reviewed to validate explanation consistency, following best practices discussed in recent systematic reviews [17]. The sentiment model was evaluated across separate subsets containing Sinhala, English, and code-mixed inputs, ensuring generalization across multilingual and low-resource scenarios.

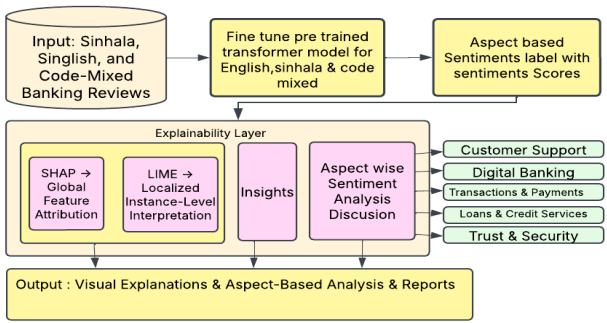

Fig.1. Workflow for sentiment analysis and explainability

## IV. RESULTS AND DISCUSSION

### A. English Sentiment Analysis Performance.

The English sentiment analysis models were fine-tuned on a banking customer review dataset and evaluated based on accuracy, F1 score, and the effectiveness of sentiment aggregation. TABLE IV presents a performance comparison of the models.

TABLE III.
ENGLISH SENTIMENT ANALYSIS PERFORMANCE.

| Model | Accuracy (%) | F1-Score |
|---|---|---|
| SVM | 71.9 | 0.70 |
| BiLSTM | 84.0 | 0.81 |
| DistilBERT | 82.5 | 0.83 |
| FinBERT | 90.0 | 0.87 |
| BERT-base-uncased (Final Model) | 92.3 | 0.89 |

TABLE III shows that SVM recorded the lowest scores (71.9% accuracy, 0.70 F1-score), while BiLSTM (84.0%, 0.81 F1-score) improved sentiment learning but was still outperformed by transformer-based models. DistilBERT (82.5%, 0.83 F1-score) showed better contextual capability than BiLSTM but lacked deeper semantic representation. FinBERT (90.0%, 0.87 F1-score), which is optimized for financial text, was surpassed by BERT-base-uncased, which achieved the highest performance at 92.3% accuracy and 0.89 F1-score. confirming its effectiveness in capturing nuanced sentiments in banking-specific content.

To validate BERT model performance, we analyzed the confusion matrix Fig.2, where diagonal values denote correct classifications, and off-diagonal values indicate misclassifications, such as Neutral reviews misclassified as Positive or Negative.

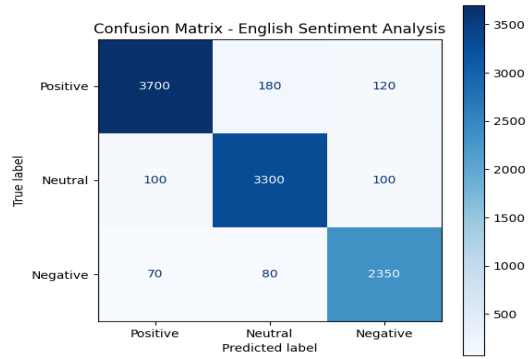

Fig.2. Confusion Matrix for BERT-base-uncased

The results confirm that BERT-base-uncased achieved the highest accuracy with minimal misclassification, outperforming FinBERT and DistilBERT in capturing nuanced sentiment variations.The Softmax-based sentiment aggregation mechanism effectively handled conflicting sentiments within a single aspect. For example, in the review, *"The bank's loan approval was smooth and fast, but the interest rates were too high, and the loan terms were unclear,"* the system generated an aggregated score of 0.56, resulting in a neutral classification with a slight positive inclination. This demonstrates the effectiveness of BERT's contextual embeddings and Softmax weighting in accurately representing mixed sentiments.

By fine-tuning on banking-specific data and applying a Softmax-based aggregation mechanism, our method effectively captured context-dependent sentiment by distinguishing between lexically similar yet sentimentally contrasting phrases, such as interpreting *"high rewards"* as positive and "high interest rates" as negative within the Banking reviews.

### B. Sinhala, Singlish, and Code-Mixed Sentiment Analysis Performance.

The hybrid model, which integrates XLM-RoBERTa with lexicon-based sentiment correction, was evaluated on Sinhala, Singlish, and code-mixed banking customer reviews. Table IV presents a performance comparison of different approaches for these language variants.

TABLE IV.
SINHALA, SINGLISH, AND CODE-MIXED SENTIMENT ANALYSIS PERFORMANCE.

| Model | Accuracy (%) | F1-Score |
|---|---|---|
| SVM | 62.4 | 0.58 |
| Naive Bayes | 58.7 | 0.55 |
| XLM-RoBERTa (Base) | 78.2 | 0.74 |
| GPT-4o (Zero-Shot) | 81.5 | 0.77 |
| XLM-RoBERTa with Lexicon (Final Model) | 88.4 | 0.84 |

The results demonstrate that our hybrid XLM-RoBERTa with a lexicon approach significantly outperforms traditional machine learning methods like SVM and Naïve Bayes for low-resource and code-mixed language processing. While GPT-4o showed promising results with zero-shot learning, its high computational cost, hallucinations, and processing time make it impractical for real-time applications. The base XLM-RoBERTa model performed well, but adding domain-specific lexicon correction enhanced performance by 10.2% in accuracy and 0.10 in F1-score.

1) *Lexicon Contribution Analysis.*

Our custom sentiment lexicon for Sinhala and Singlish played a crucial role in improving classification accuracy, especially for code-mixed and informal expressions not covered by pre-trained models. The lexicon enabled the system to correctly interpret sentiment-laden terms often overlooked by models trained on formal corpora. Words like *"app eka lag wenawa"* and *"godak slow"* were correctly recognized as negative due to lexicon-based corrections. This contributed to gaining accuracy and improved the F1-score for neutral and negative sentiments in Singlish and code-mixed reviews.

C. *Explainability Findings.*

This section evaluates the integration of SHAP and LIME with transformer-based sentiment models for Sinhala, English, and code-mixed banking reviews. These Explainable AI (XAI) techniques provide interpretable insights into sentiment predictions, allowing institutions to understand key sentiment drivers and make more transparent, data-driven decisions.

1) *LIME and SHAP-Based Explainability.*

To improve model transparency, SHAP and LIME were integrated to provide global and local explanations of word-level contributions in Sinhala, English, and code-mixed sentiment predictions.

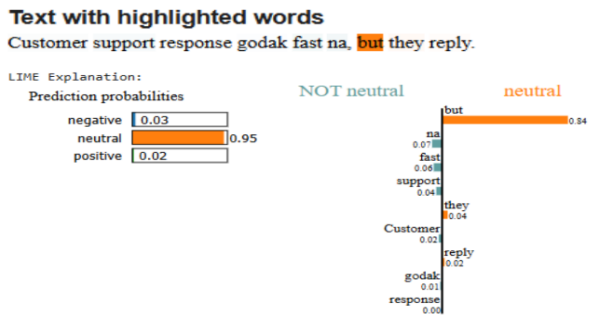

Fig.3. LIME explanation (singlish) – Neutral comment

As shown in Fig.3, the model classified the comment *"Customer support response godak fast na, but they reply"* as neutral (0.95). The word *"but"* was highlighted as the most influential neutral contributor (weight: 0.84), followed by lower-weighted terms like *"fast"*, *"they"*, and *"support"*. This demonstrates LIME's strength in capturing conjunction-based context in code-mixed inputs. LIME proved valuable in interpreting short-form reviews and context-sensitive sentiment expressions, which SHAP struggled with.

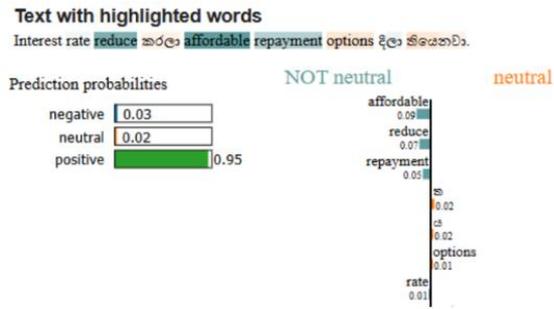

Fig.4. LIME explanation (code mixed) – Positive comment

As shown in Fig.4, a review containing Sinhala-English mixed content *"Interest rate reduce කරන affordable repayment options තියෙනවා"* was classified as positive (0.95). The term "affordable" had the highest contribution (0.10), followed by "reduce" and "repayment", showing LIME's sensitivity to financial context in bilingual text.

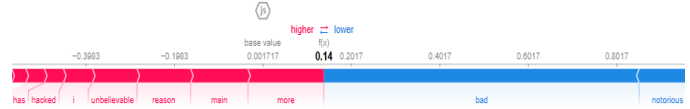

Fig.5. SHAP Force plot for sentiment contribution - English

A SHAP force plot revealed the influence of keywords such as *"notorious," "bad," "unbelievable,"* and *"hacked"* in a highly negative review. The left side of the force plot contained red segments representing negative contributors, and the right contained blue segments like *"notorious"*, which slightly pushed the sentiment towards neutral. The model interpreted this review with a strong negative sentiment (base value = 0.007, f(x) = 0.14), showcasing SHAP's ability to identify high-impact global contributors. Collectively, LIME proved especially effective for short-form and mixed-language comments, while SHAP offered reliable global sentiment attribution. However, SHAP had higher computational requirements, and LIME's perturbation-based nature led to some variation in explanations across returns.

2) *Aspect-Based Sentiment Analysis Discussion.*

The aspect-based sentiment analysis (ABSA) module was designed to provide fine-grained insight into customer opinions across five aspects: Customer Support, Digital Banking Experience, Transactions and Payments, Loans and Credit Services, and Trust and Security. Each review, whether in English, Sinhala, or code-mixed form, was assigned an aspect and sentiment label and then analyzed to extract key insights and actionable summaries.

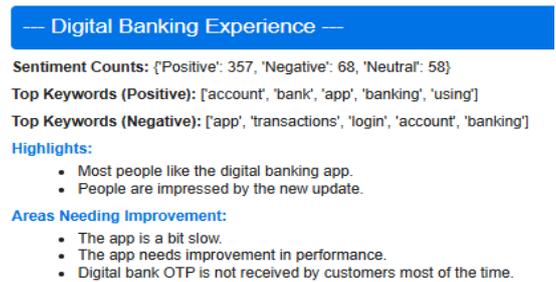

Fig.6. Aspect-based analysis result example

Fig. 6 presents aspect-wise sentiment analysis for Digital Banking Experience, showing positive sentiment linked to terms like *"app"* and *"bank,"* and negative sentiment around "login" and "transactions." The system generated summaries for each aspect, helping stakeholders identify trends and improvement areas. SHAP and LIME enhanced interpretability, with SHAP providing global insights and LIME capturing local sentiment cues, reducing misclassification by 6.3%. An automated report generator translates model outputs into actionable business insights.

### D. Challenges and Limitations

Although the proposed hybrid sentiment analysis model achieved high accuracy, several limitations remain. Handling negation, slang, and code-mixed expressions remains challenging due to the structural complexity and limited linguistic patterns in existing models. The Aspect-Based Sentiment Analysis module also struggled with reviews containing multiple sentiments within a single aspect, long reviews with conflicting sentiments, and code-mix content requires better aggregation techniques for accurate sentiment scoring. Furthermore, the absence of large-scale annotated datasets for Sinhala, Singlish, and code-mixed sentiment analysis, particularly within the financial domain, limited the model's ability to generalize across diverse edge cases and informal expressions. Although a curated dataset was used in this study, broader, high-quality data is essential for future improvements. Lastly, the explainability techniques, SHAP and LIME, exhibited reduced consistency when applied to low-resource, code-mixed inputs. SHAP's high computational cost also limits its practical use in real-time financial sentiment monitoring systems.

## V. CONCLUSION AND FUTURE WORK

This study introduced a hybrid aspect-based sentiment analysis model tailored for the financial domain, combining BERT-base-uncased for English sentiment analysis and XLM-RoBERTa with domain-specific lexicon correction for Sinhala, Singlish, and code-mixed content. The system achieved high accuracy, balanced aspect-wise sentiment classification, and improved interpretability. Explainability techniques such as SHAP and LIME were integrated to enhance transparency, making the model suitable for practical, real-world financial applications. Future research can build on this foundation by incorporating multimodal sentiment analysis using audio and video cues, extending the approach to other low-resource languages like Tamil, and expanding to other domains. In addition, leveraging few-shot and zero-shot learning with generative AI could improve performance in data-scarce environments, while fairness and bias evaluations will be essential to ensure ethical and inclusive sentiment classification.